\documentclass[10pt,twocolumn,letterpaper]{article}

\usepackage{iccv}
\usepackage{times}
\usepackage{epsfig}
\usepackage{graphicx}
\usepackage{xcolor}
\usepackage{amsmath}
\usepackage{amssymb}
\usepackage{listings}
\usepackage{booktabs}

\usepackage[breaklinks=true,bookmarks=false]{hyperref}

\iccvfinalcopy 

\def\iccvPaperID{****} 

\ificcvfinal\fi

\begin{document}

\title{Guiding Video Prediction with Explicit Procedural Knowledge}

\author{Patrick Takenaka$^{1,2}$, Johannes Maucher$^1$, Marco F. Huber$^{2,3}$\\
Institute for Applied AI, Hochschule der Medien Stuttgart, Germany$^1$\\
Institute of Industrial Manufacturing and Management IFF, University of Stuttgart, Germany$^2$\\
Fraunhofer Institute for Manufacturing Engineering and Automation IPA, Stuttgart, Germany$^3$\\
{\tt\small \{takenaka,maucher\}@hdm-stuttgart.de, marco.huber@ieee.org}
}

\maketitle
\ificcvfinal\thispagestyle{empty}\fi

\begin{abstract}
We propose a general way to integrate procedural knowledge of a domain into deep learning models. We apply it to the case of video prediction, building on top of object-centric deep models and show that this leads to a better performance than using data-driven models alone. We develop an architecture that facilitates latent space disentanglement in order to use the integrated procedural knowledge, and establish a setup that allows the model to learn the procedural interface in the latent space using the downstream task of video prediction. We contrast the performance to a state-of-the-art data-driven approach and show that problems where purely data-driven approaches struggle can be handled by using knowledge about the domain, providing an alternative to simply collecting more data.
\end{abstract}

\section{Introduction}

The integration of expert knowledge in deep learning systems reduces the complexity of the overall learning problem, while offering domain experts an avenue to add their knowledge into the system, potentially leading to improved data efficiency, controllability, and interpretability. \cite{vonruedenInformedMachineLearning2023} showed in detail the various types of knowledge that are currently being integrated in deep learning, ranging from logic rules to regularize the learning process \cite{xuSemanticLossFunction2018}, to modelling the underlying graph structure in the architecture \cite{wattersVisualInteractionNetworks2017a}. Especially in the physical sciences, where exact and rigid performance of the model is of importance, data-driven systems have shown to struggle on their own, and many complex problems cannot be described through numerical solvers alone \cite{wangPhysicsGuidedDeepLearning2023}. This highlights the need for hybrid modelling approaches that make use of both theoretical domain knowledge, and of collected data. When viewing this approach from the perspective of deep learning, if the model is able to understand and work with integrated domain knowledge, it could potentially render many data samples redundant w.r.t. information gain. 

In addition to the recognized knowledge integration categories \cite{vonruedenInformedMachineLearning2023}, we propose to view procedural knowledge described through programmatic functions as its own category, as it is equally able to convey domain information in a structured manner as other types, while bringing with it an already established ecosystem of definitions, frameworks, and tools. Such inductive domain biases in general can help models to obtain a more structured view of the environment~\cite{goyalInductiveBiasesDeep2022} and lead them towards more desirable predictions by either restricting the model hypothesis space, or by guiding the optimization process~\cite{borghesiImprovingDeepLearning2020}.

We argue that by incorporating procedural knowledge we can give neural networks powerful learning shortcuts where data-driven approaches struggle, and as a result reduce the demand for data, allow better out-of-distribution performance, and enable domain experts to control and better interpret the predictions. In summary, our contributions are:
\begin{itemize}
    \item Specification of a general architectural scheme for procedural knowledge integration.
    \item Application of this scheme to video prediction, involving a novel latent space separation scheme to facilitate learning of the procedural interface.
    \item Performance analysis of our proposed method in contrast to a purely data-driven approach.
\end{itemize}

The paper is structured as follows: First, our proposed procedural knowledge integration scheme is introduced in Sec.~\ref{sec:arch}, followed by its specification for the video prediction use case in Sec.~\ref{sec:vp}. We show relevant related work in Sec.~\ref{sec:related} and continue by describing the concrete model and overall setup that we used in Sec.~\ref{sec:setup}, after which several experiments regarding the model performance and feasibility are made in Sec.~\ref{sec:exp}.

\section{Proposed Architecture}\label{sec:arch}
We view the integrated procedural knowledge as an individual module in the overall architecture, and the learning objective corresponds to the correct utilization of this module, i.e., the learning of the program interface, to solve the task at hand. More specifically, we consider the case where the integrated knowledge is only solving an intermediate part of the overall task, i.e., it neither directly operates on the input data, nor are its outputs used as a prediction target.

\begin{figure}
    \centering
    \includegraphics[scale=.8]{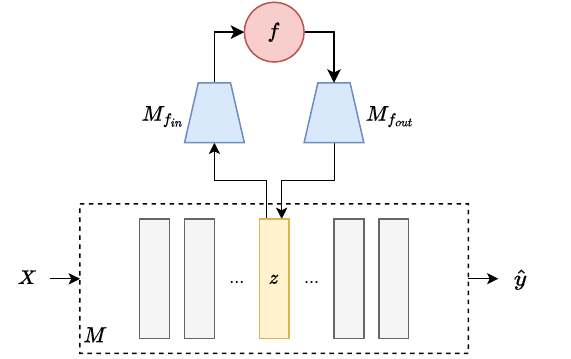}
    \caption{Abstract structure of our proposed procedural knowledge integration interface. Features of $X$ are extracted in model $M$, resulting in intermediate feature maps (shown in grey). Out of these, a selected feature map $z$ is then used to decode it into the input space of the integrated procedural module $f$ through $M_{f_\mathrm{in}}$, and the output of $f$ is encoded back into the latent space of $z$ using $M_{f_\mathrm{out}}$. $M$ continues with this updated latent state to obtain prediction $\hat{y}$.}
    \label{fig:abstract_integration}
\end{figure}

More formally, given data sample $X$ and procedural module $f$, the model latent state $z$ is decoded into and encoded from the function input space through learned modules $M_{f_\mathrm{in}}$ and $M_{f_\mathrm{out}}$, respectively. Here, $z$ corresponds to an intermediate feature map of an arbitrary deep learning model $M$ whose target domain at least partially involves processes that are described in $f$. The output of $M_{f_\mathrm{out}}$ is then fused with $z$ using an arbitrary operator $\oplus$. This structure is shown in Fig.~\ref{fig:abstract_integration}. 

Procedural knowledge in general and programmatic functions in particular operate on a discrete set of input and output parameters. The aforementioned interface thus needs to disentangle the relevant parameters in the distributed representation and bind them to the correct inputs, and perform the reverse operation on the output side, tasks that are still challenging in many cases \cite{greffBindingProblemArtificial2020}.

We show that in our setup we are able to learn this interface implicitly by focusing on a downstream task instead.

\subsection{Case Study: Video Prediction}\label{sec:vp}
Video Prediction is an important objective in the current deep learning landscape. With it, many visual downstream tasks can be enhanced or even enabled that utilize temporal information. Example tasks are model predictive control (MPC) \cite{jaquesPhysicsasInverseGraphicsUnsupervisedPhysical2019}, visual question answering (VQA) \cite{wuSlotFormerUnsupervisedVisual2023}, system identification \cite{jaquesPhysicsasInverseGraphicsUnsupervisedPhysical2019}, or even content generation \cite{huMakeItMove2022}. 

Some of these benefit even more if the system is controllable and thus, allows the integration of human intention into the inference process. This is typically done by conditioning the model on additional modalities such as natural language \cite{huMakeItMove2022,xuControllableVideoGeneration2023} or by disentangling the latent space \cite{tulyakovMoCoGANDecomposingMotion2018,wangG3ANDisentanglingAppearance2020}.

More recently, researchers have shown~\cite{wuSlotFormerUnsupervisedVisual2023} that object-centric learning offers a suitable basis for video prediction, as learning object interactions is difficult without suitable representations. We propose to build on top of such models, since knowledge about objects in the environment is an integral aspect of many domain processes and as such facilitates our approach.

We proceed by reducing these distributed object-centric representations further to individual object properties, which are then usable by our procedural module---a simple differentiable physics engine modeling the underlying scene dynamics.

We follow the approach of SlotFormer \cite{wuSlotFormerUnsupervisedVisual2023} and utilize a frozen pretrained Slot Attention for Video (SAVi) \cite{kipfConditionalObjectCentricLearning2022} model trained on video object segmentation to encode and decode object latent states for each frame. Also similarly, our proposed model predicts future frames in an auto-regressive manner using a specialized rollout module, with the assumption that the first $N$ frames of a video are given in order to allow the model to observe the initial dynamics of the scene.

\begin{figure}[tb]
    \centering
    \includegraphics[scale=.65]{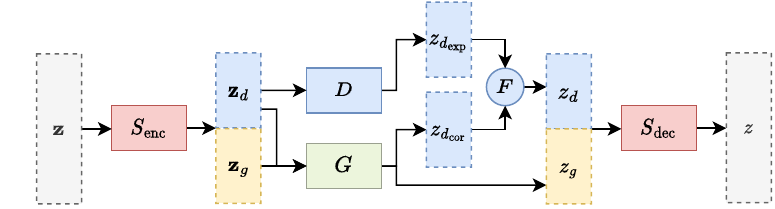}
    \caption{Overview of the prediction of latent state $z$ at time step $t$, given the previous latent states $\textbf{z}$ of time steps $t-N\ldots t-1$, where $N$ is the number of context frames. The encoder model $S_{\mathrm{enc}}$ transforms the fixed latent space of $\textbf{z}$ into a separable latent space composed of dynamics state $\textbf{z}_d$ and Gestalt state $\textbf{z}_g$. Both are fed through the joint Gestalt and dynamics prediction model $G$ to obtain dynamics correction $z_{d_{\mathrm{cor}}}$ and future Gestalt state $z_g$, whereas only $\textbf{z}_d$ is given to the explicit dynamics model $D$ to get the explicit dynamics prediction $z_{d_{\mathrm{exp}}}$. Both $z_{d_{\mathrm{exp}}}$ and $z_{d_{\mathrm{cor}}}$ are fused with fusion method $F$, resulting in the future dynamics state $z_d$. Finally, $z_d$ and $z_g$ are concatenated and fed through the state decoder $S_{\mathrm{dec}}$ to obtain future latent state $z$. In terms of Fig. \ref{fig:abstract_integration}, the integrated physics engine within $D$ corresponds to $f$, with $M_{f_{\mathrm{in}}}$ being the computational graph starting from $S_{\mathrm{enc}}$ up until the physics engine, and $M_{f_{\mathrm{out}}}$ the subsequent dynamics computations until after $S_{\mathrm{dec}}$. The dynamics correction and Gestalt computations are not shown explicitly and are part of $M$.}
    \label{fig:rollout_arch}
\end{figure}

Within the rollout module, our first goal for each object latent state is to disentangle object factors that are relevant as function input from those that are not. In our case, these are the dynamics and appearance---or Gestalt \cite{traubLearningWhatWhere2023}---factors, respectively. However, as the upstream SAVi model is frozen and did not assume such disentanglement, we first have to apply a non-linear transformation on its latent space to enable the model to learn to separate the latent state into dynamics and Gestalt parts based on the inductive biases of the architecture that follows. We then use these---still distributed---latent states to obtain discrete physical state representations---i.e., in our case 3D vectors representing position and velocity---that can be processed by our explicit dynamics module in order to predict the state of the next time step. In order to avoid bottlenecks in the information flow, we introduce a parallel model that predicts both a dynamics correction and the future Gestalt state. The reasoning here is that in many cases both are dependent of each other and thus, need to be modelled jointly. Both dynamics predictions are then averaged over to produce the final dynamics state. The fused dynamics state and the predicted Gestalt state are finally concatenated to obtain the latent state of the next time step. This latent state is finally transformed non-linearly back into the latent space of the pretrained SAVi model, before it is decoded into pixel space. The rollout module can be seen in detail in Fig.~\ref{fig:rollout_arch}. We verify in our experiments that even without additional auxiliary loss terms to regularize the latent state our model is able to correctly utilize the integrated dynamics module, indicating that the inductive bias of a correctly predicted and decoded physics state is sufficient for better visual predictions.

\section{Related Work}\label{sec:related}
\textbf{Physics-Guided Deep Learning for Videos.} The explicit representation of dynamics prevalent in a video within a deep learning model is a popular shortcut to learning the underlying concepts in the scene, and oftentimes necessary due to the inherent difficulty and ambiguity of many tasks~\cite{wangPhysicsGuidedDeepLearning2023}. The main objectives are usually the estimation of underlying system parameters and rules \cite{wuPhysics101Learning2016,xuBayesianSymbolicApproachReasoning2021,linkPredictingPhysicalObject2022}, or the adherence of the model output to certain environmental constraints \cite{bezenacDeepLearningPhysical2019,yinAugmentingPhysicalModels2021,wuPastNetIntroducingPhysical2023,yinContinuousPDEDynamics2022}, leading to more accurate predictions. With that---as is the case for neuro-symbolic approaches~\cite{wuGalileoPerceivingPhysical2015,wuPhysics101Learning2016,yuSurveyNeuralsymbolicSystems2021} in most cases---the idea is to also inherently benefit from an improvement in interpretability and data efficiency.

A long-standing approach is to represent the dynamics by an individual module, i.e., a physics engine, and use different means to join it with a learnable model. Early work \cite{wuGalileoPerceivingPhysical2015,wuPhysics101Learning2016} utilized this to predict physical outcomes, while simultaneously learning underlying physical parameters. Later work extended this towards video prediction, in which the output of the physics engine is used for rendering through a learnable decoder. Some used custom decoder networks for the given task \cite{jannerReasoningPhysicalInteractions2018,yangLearningPhysicsConstrained2022}, or integrated a complete differentiable renderer in addition \cite{murthyGradSimDifferentiableSimulation2020}. However, these were limited to specialized use cases for the first, and required perfect knowledge of the visual composition of the environment for the latter. Another common direction is the use of Spatial Transformers (STs) \cite{NIPS2015_33ceb07b}, since they allow easy integration of spatial concepts such as position and rotation in the decoding process. However, these approaches \cite{kosiorekSequentialAttendInfer2018a,jaquesPhysicsasInverseGraphicsUnsupervisedPhysical2019,kandukuriPhysicalRepresentationLearning2022}---albeit similar to our approach---assumed that (1) no data-driven correction of physics state is necessary and (2) the visuals of the scene outside of the dynamics properties remain static and can be encoded in the network weights, limiting their applicability to more complex settings. With our proposed approach we can model such properties.

For object-centric scenarios it is common to also take into account the relational structure of dynamical scenes in order to model object interactions by utilizing graph-based methods in the architecture \cite{battagliaInteractionNetworksLearning2016a,bearLearningPhysicalGraph2020,kossenStructuredObjectAwarePhysics2019,tangIntrinsicPhysicalConcepts2023}.

\textbf{Disentangled Video Dynamics.} Latent factor disentanglement in general assumes that the data is composed of a set of---sometimes independent---latent factors. Once the target factors can be disentangled, control over the environment becomes possible, and as such these approaches are of special interest in generative models. Early work heavily built on top of Variational Autoencoders (VAEs) \cite{higginsBetaVAELearningBasic2022}. However, later on it was proven that inductive biases are necessary to achieve disentanglement, and earlier work instead only exploited biases in the data \cite{locatelloSoberLookUnsupervised2020}. Typically, these inductive biases are in the form of factor labels \cite{locatelloDisentanglingFactorsVariations2019}. Such models were also used for disentanglement of physical properties and dynamics \cite{zhuLearningDisentangleLatent2019,thoreauVAEPhysicsintegratedGenerative2023}. In this domain, instead of only providing labels to achieve disentanglement, it is also common to help the model discover underlying dynamics by modeling them as Partial Differential Equations (PDEs)\cite{leguenDisentanglingPhysicalDynamics2020,donaPDEDrivenSpatiotemporalDisentanglement2021,yinAugmentingPhysicalModels2021}. For video data that does not necessarily follow certain physical rules, some use a more general approach and focus on the disentanglement of position and Gestalt factors, with the idea that many object factors are independent of their position in the frame \cite{traubLearningWhatWhere2023}. Having explicit encoding or decoding processes also helps in obtaining disentangled dynamics \cite{kosiorekSequentialAttendInfer2018a,jaquesPhysicsasInverseGraphicsUnsupervisedPhysical2019,murthyGradSimDifferentiableSimulation2020,kandukuriPhysicalRepresentationLearning2022}. 

\section{Setup}\label{sec:setup}
As is done in the original SAVi paper \cite{kipfConditionalObjectCentricLearning2022}, we condition the SAVi slots on the first frame object bounding boxes and pre-train on sequences of six video frames, optimizing the reconstruction of the optical flow map for each frame. Experiments have shown that optical flow reconstruction leads to better object segmentations, which we find is a better proxy for evaluating correct object dynamics than video reconstruction itself. After convergence we freeze the SAVi model.

For the video prediction task, we encode the initial six frames using this frozen model, and use these as initial context information for the video prediction model. We then let the model auto-regressively predict the next 12 frames during training---or 24 frames during validation---always keeping the most recent six frames as reference. While more than a single reference frame would not be necessary for the integrated dynamics knowledge, the six frames are instead used in the transformer-based joint dynamics and Gestalt predictor model. In order to give the model a hint about the magnitude of the dynamics state values, we condition the dynamics state of the first frame on the ground-truth state.

\subsection{Implementation Details}
For the SAVi model we mainly follow the implementations of SlotFormer \cite{wuSlotFormerUnsupervisedVisual2023} and the original work \cite{kipfConditionalObjectCentricLearning2022}. The encoder consists of a standard Convolutional Neural Network (CNN) with a subsequent positional embedding. To obtain slot representations for a given frame we perform two iterations of slot attention, followed by a transformer model with multi-head self attention for modelling slot interactions and a final Long Short-Term Memory (LSTM) model in order to transition the representation into the next time step. We set the number of slots to six, each with size 128. The representations obtained after the slot attention rounds are decoded into the target frames using a Spatial Broadcast Decoder \cite{wattersSpatialBroadcastDecoder2019b} with a broadcast size of 8.

For the video prediction model we denote the most recent $N$ context frame representations of time steps $t-N\ldots t-1$ in bold as $\textbf{z}$ and the latent representation prediction for time step $t$ as $z$ in order to improve readability. Both the latent state encoder $S_{\mathrm{enc}}$ and decoder of the video prediction model $S_{\mathrm{dec}}$ are MLPs, each with a single ReLU activated hidden layer of size 128. They have shown to introduce sufficient non-linearity to allow state disentanglement. The latent state obtained from $S_{\mathrm{enc}}$ is kept the same size as the slot size and is split into two equally sized parts $\textbf{z}_d$ and $\textbf{z}_g$ for the subsequent dynamics and Gestalt models.

The dynamics model---i.e., the explicit physics engine---takes a physical state representation consisting of a 3D position and 3D velocity of a single frame as input, which is obtained from a linear readout layer of the most recent context frame of latent state $\textbf{z}_d$, or directly from groundtruth for the very first predicted frame. The physics engine itself is fully differentiable and consists of no learnable parameters. It calculates the dynamics taking place as in the original data simulation using a regular semi-implicit Euler integration scheme. Pseudo code of this engine can be seen in Listing \ref{lst:integrated_function}. Its output---consisting of again a 3D position and 3D velocity of the next timestep---is then transformed back into the latent state $z_{d_{\mathrm{exp}}}$ with another linear layer. For the Gestalt properties we utilize a prediction setup and configuration as in the original SlotFormer model: First, the latent state $\textbf{z}_g$ is enriched with temporal positional encodings after which a multi head self attention transformer is used for obtaining future latent representations $z_{d_{\mathrm{cor}}}$ and $z_g$. Both $z_{d_{\mathrm{exp}}}$ and $z_{d_{\mathrm{cor}}}$ are merged by taking their mean, and the resulting vector $z_{d}$ is concatenated with $z_g$ in order to obtain the latent representation of the future frame. $S_{\mathrm{dec}}$ is finally used to transform this vector back into the latent representation $z$ of SAVi, where it can be decoded into pixel space by the pretrained frozen SAVi decoder.

\lstset{language=Python,commentstyle=\color[HTML]{228B22}\sffamily, basicstyle=\scriptsize\sffamily, emph={[3]range,sum,pow,sqrt}, emphstyle={[3]\color[HTML]{D17032}},
  emph={[4]for,in,def,get_pos_delta}, emphstyle={[4]\color{blue}},}
\begin{lstlisting}[basicstyle=\small\sffamily, label={lst:integrated_function}, language=Python, caption=Python pseudo code of the integrated function for our data domain which calculates a future physical state consisting of position and velocity of each object. $G$ in the code corresponds to the gravitational constant. As is done in the original simulation each predicted frame is subdivided into smaller simulation steps---a standard approach for numerical-based physics simulations.]
def dynamics_step(pos, vel):
  for sim_idx in range(simulation_steps):
    # Position delta between objects
    pos_delta = get_pos_delta(pos)
    # Squared distances between objects
    r2 = sum(pow(pos_delta, 2))
    # Calculate force direction vector
    F_dir = pos_delta / sqrt(r2)
    # Calculate force
    F = F_dir * (G * (mass / r2))
    # F = ma
    a = F / mass
    # Semi-implicit euler
    vel = vel + simulation_dt * a
    pos = pos + simulation_dt * vel
  return pos, vel
\end{lstlisting}

\subsection{Data}
Our dataset consists of a simulated environment of multiple interacting objects resulting in complex nonlinear dynamics. The idea was to generate an object-centric dataset for which current state-of-the-art video prediction models struggle and where the integration of knowledge about the environment is possible and sensitive. Datasets used in existing object-centric video prediction literature either did not feature complex nonlinear dynamics, or involved non-differentiable dynamics (\eg, collisions) that are out of scope for now. However for the latter we note that non-differentiable dynamics such as collisions could still be integrated with our approach by building a computational graph that covers all conditional pathways. Although this approach is computationally more inefficient and does not directly convey collision event information to the learning algorithm, work exists \cite{dingDynamicVisualReasoning2021a} that show that this can still be exploited well enough and is simultaneously easy to implement in current deep learning frameworks with dynamic computational graphs.

The future states are predicted using a simple physics engine that simulates gravitational pull between differently sized spherical objects without collisions, as in the three body problem \cite{musielakThreebodyProblem2014}. In order to keep objects in the scene, we add an invisible gravitational pull towards the camera focus point and limit the movement in $x$ and $y$ direction. Objects are then rendered in 3D space using slight illumination and no background. Each object can have different material properties, which change their visuals slightly.

We create 10k RGB video samples consisting of 32 frames and spatial size $64\times64$ each with their corresponding optical flow and segmentation masks using kubric \cite{greffKubricScalableDataset2022}, which combines a physical simulator with a 3D rendering engine, allowing the generation of arbitrary physical scenes. We render four frames per second, and subdivide each frame into 60 physical simulation steps. Each sample uses the same underlying dynamics but with different starting conditions for the objects. The number of objects randomly varies per sample from 3-5 objects. For each object, we also store its physical state at each frame consisting of the 3D world position and velocity. All objects have the same fixed mass.

\section{Experiments}\label{sec:exp}
In all experiments, we compare our proposed architecture with a  SlotFormer model, representing a purely data-driven approach. To improve comparability, the transformer architectures of both our joint dynamics and Gestalt predictor $G$ and the SlotFormer rollout module are the same. Also, both use the same underlying frozen SAVi model as object-centric encoder and decoder.

We train SAVi and the video prediction models for at maximum 100k steps each or until convergence is observed by early stopping, using a batch size of 64. We clip gradients to a maximum norm of $0.05$ and train using Adam with an initial learning rate of $0.0001$. 

For evaluation purposes, we report the aggregated object segmentation performance over three seeds using the Adjusted Rand Index (ARI) and mean Intersection-Over-Union (mIoU) scores, in addition to their foreground (FG) variants ARI-FG and mIoU-FG which disregard background predictions.

We first analyze the baseline performance of our proposed approach in Sec.~\ref{sec:exp_baseline}, followed by an experiment focusing on the completeness of the integrated function in Sec.~\ref{sec:exp_inaccurate}. We then consider the model performance for very limited data availability in Sec.\ref{sec:exp_data} and conclude with an ablation experiment regarding the latent state separation in Sec.\ref{sec:exp_joint}.

\subsection{Baseline}\label{sec:exp_baseline}
Here, we integrate the complete underlying dynamics of the environment in our model. As such, we also verify the utility of still keeping a parallel auto-regressive joint Gestalt and dynamics model by replacing it with an identity function and observing the performance, since with perfect knowledge about the dynamics and the initial frame appearance the model should have all necessary information for an accurate prediction.

\begin{table}[htb]
    \centering
    \caption{Performance comparison of a purely data-driven model (SlotFormer), our proposed model (Ours), and a variant of our architecture with an identity function as the joint Gestalt and dynamics predictor (Ours-Pure). For reference the performance of the underlying SAVi model is also reported, describing the upper bound performance that any downstream video prediction model can achieve.}
    \label{tab:base_performance}
    \resizebox{.48\textwidth}{!}{%
    \begin{tabular}{l|l|l|l|l}
       &                         \textbf{mIoU}$\uparrow$ &                      \textbf{mIoU-FG}$\uparrow$ &                          \textbf{ARI}$\uparrow$ &                       \textbf{ARI-FG}$\uparrow$ \\
\midrule
    \textbf{Ours} & $\textbf{32.1}$\textcolor{darkgray}{\scriptsize$\pm0.7$} & $\textbf{29.2}$\textcolor{darkgray}{\scriptsize$\pm0.8$} & $\textbf{48.2}$\textcolor{darkgray}{\scriptsize$\pm1.0$} & $\textbf{82.9}$\textcolor{darkgray}{\scriptsize$\pm2.8$} \\
 Ours-Pure & $29.5$\textcolor{darkgray}{\scriptsize$\pm0.1$} & $26.1$\textcolor{darkgray}{\scriptsize$\pm0.2$} & $43.8$\textcolor{darkgray}{\scriptsize$\pm0.2$} & $71.6$\textcolor{darkgray}{\scriptsize$\pm0.6$} \\
SlotFormer & $15.1$\textcolor{darkgray}{\scriptsize$\pm0.1$} &  $9.4$\textcolor{darkgray}{\scriptsize$\pm0.1$} & $16.9$\textcolor{darkgray}{\scriptsize$\pm0.3$} &  $6.1$\textcolor{darkgray}{\scriptsize$\pm0.4$} \\
\midrule

         SAVi & 36.1 & 34.0 & 55.1 & 93.0
    \end{tabular}}
\end{table}

As we can see in Tab.~\ref{tab:base_performance}, our proposed architecture outperforms a purely data-driven approach such as SlotFormer by a large margin, and comes close to the performance of the underlying SAVi model, which in contrast to video prediction methods has access to every video frame and simply needs to segment them. However, even when integrating perfect dynamics knowledge it is still beneficial to keep a parallel data-driven Gestalt and dynamics predictor, highlighting the need to model the dependency between appearance and dynamics in the scene. Both our models are also able to predict the future object positions and velocities in the physics state space accurately, with a Mean Absolute Error (MAE) close to 0 across all predicted frames when compared to the groundtruth.

\begin{figure}[t]
    \centering
    \includegraphics[scale=.5]{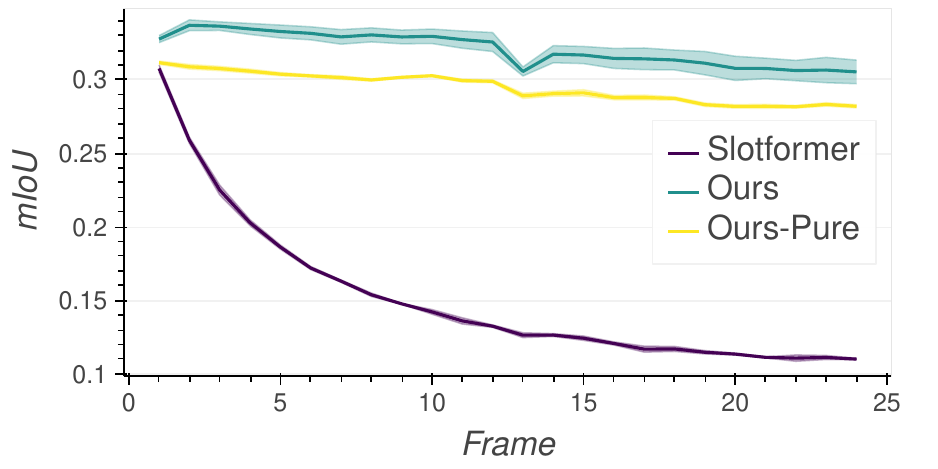}
    \caption{The mIoU performance w.r.t. each auto-regressive frame prediction. While the data-driven model exponentially becomes more inaccurate over time, the integration of the dynamics knowledge helps to keep the prediction performance stable. The pure variant of our architecture without a data-driven Gestalt and dynamics predictor follows the slope of our main architecture, albeit at a lower magnitude. Their difference indicates the missing handling of Gestalt and dynamics interdependencies.}
    \label{fig:unroll_performance}
\end{figure}

\begin{figure}[tb]
    \centering
    \includegraphics[scale=.2]{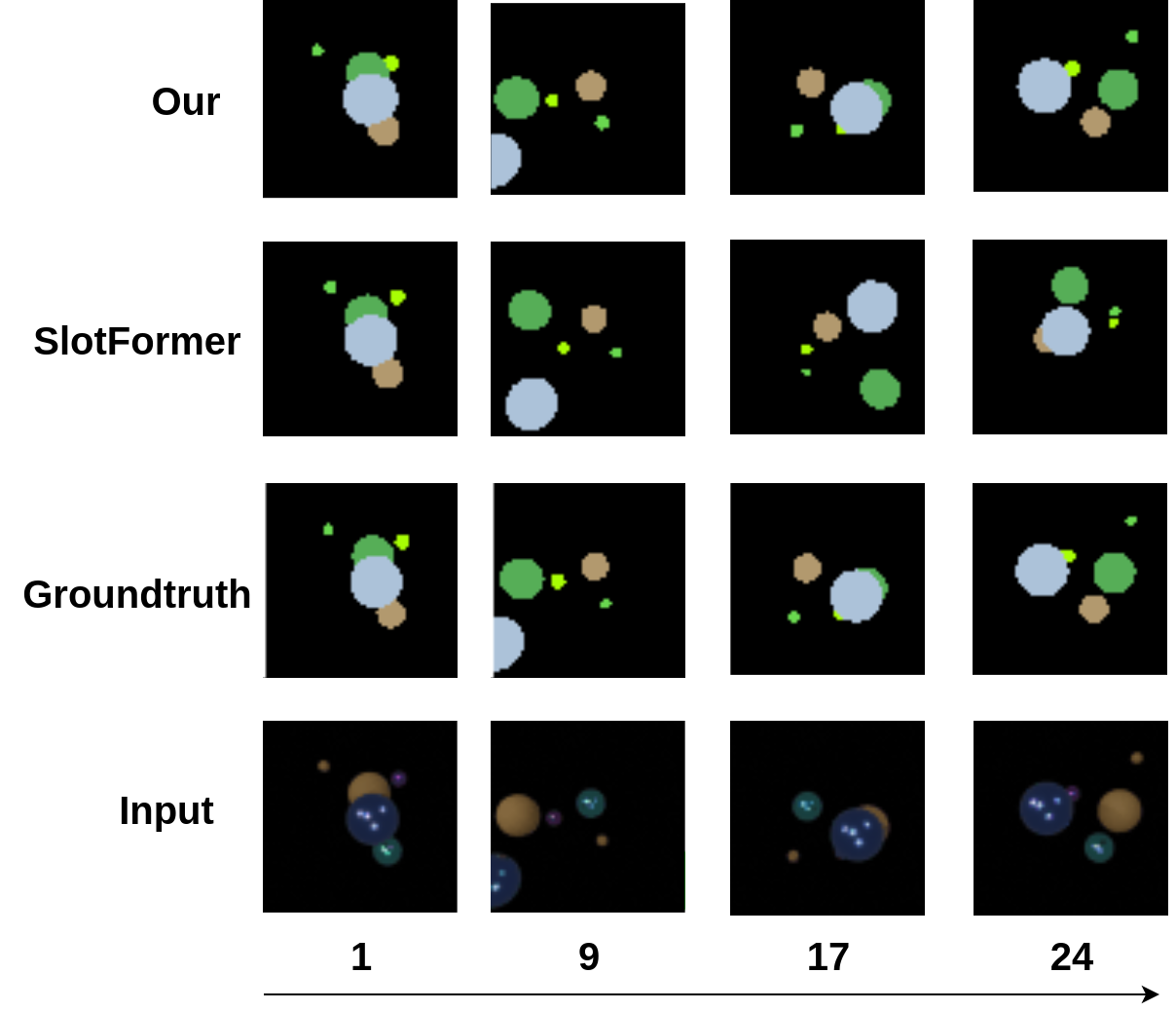}
    \caption{Sample prediction comparisons for different unroll steps. While both models are able to keep object shapes intact, the dynamics of the SlotFormer model are diverging quickly, while our model can keep up with the complex dynamics.}
    \label{fig:sample}
\end{figure}

Regarding the unroll performance, i.e., the frame-by-frame prediction performance, the SlotFormer model's performance quickly deteriorates, while both variants of our architecture keep the performance more stable over time, as seen in Fig.~\ref{fig:unroll_performance}. As seen in Fig.~\ref{fig:sample}, the performance decrease stems mainly from wrong dynamics, as the object shapes are kept intact even for the SlotFormer model.

\subsection{Inaccurate Dynamics Knowledge}\label{sec:exp_inaccurate}
In the previous setup, the integrated function described the underlying dynamics perfectly and as such might allow the model to learn undesirable shortcuts. Here, we therefore evaluate whether inaccuracies in the integrated dynamics knowledge hinder the utilization of the integrated dynamics. We introduce these inaccuracies by using wrong simulation time steps, which results in wrong state predictions, albeit with the same underlying dynamics. We report the results in Tab.~\ref{tab:imperfect_performance}.

\begin{table}[tb]
    \centering
    \caption{Performance of the model using inaccurate dynamics information (Ours-Inaccurate) in contrast to the base models. It can be observed that although the performance decreases, it still stays above that of the purely data-driven SlotFormer model.}
    \label{tab:imperfect_performance}
    \resizebox{.48\textwidth}{!}{%
    \begin{tabular}{c|c|c|c|c}
            &                         \textbf{mIoU}$\uparrow$ &                      \textbf{mIoU-FG}$\uparrow$ &                          \textbf{ARI}$\uparrow$ &                       \textbf{ARI-FG}$\uparrow$ \\
\midrule
Ours-Inaccurate & $21.7$\textcolor{darkgray}{\scriptsize$\pm0.6$} & $17.1$\textcolor{darkgray}{\scriptsize$\pm0.7$} & $29.0$\textcolor{darkgray}{\scriptsize$\pm1.4$} & $27.0$\textcolor{darkgray}{\scriptsize$\pm2.6$} \\\midrule
           Ours & $32.1$\textcolor{darkgray}{\scriptsize$\pm0.7$} & $29.2$\textcolor{darkgray}{\scriptsize$\pm0.8$} & $48.2$\textcolor{darkgray}{\scriptsize$\pm1.0$} & $82.9$\textcolor{darkgray}{\scriptsize$\pm2.8$} \\
     SlotFormer & $15.1$\textcolor{darkgray}{\scriptsize$\pm0.1$} &  $9.4$\textcolor{darkgray}{\scriptsize$\pm0.1$} & $16.9$\textcolor{darkgray}{\scriptsize$\pm0.3$} &  $6.1$\textcolor{darkgray}{\scriptsize$\pm0.4$} \\
    \end{tabular}}
\end{table}

While the performance has clearly deteriorated, it is still above the purely data-driven approach. As such we can see that just the information about the dynamics process in itself carries valuable information for the final predictions, not only the concrete dynamics state.

\subsection{Data Efficiency}\label{sec:exp_data}
Next, we analyze the prediction performance when using only 300 data samples, amounting to 3\% of the original data. We report the results in Tab.~\ref{tab:data_efficiency}. As expected, the performance of both models drops, however the SlotFormer predictions are now close to random predictions, indicated by the very low foreground scores. In contrast, our proposed model still achieves a better overall performance than the SlotFormer model using the complete dataset.

\begin{table}[tb]
    \centering
    \caption{Performance comparison of a purely data-driven model (SlotFormer) and our proposed architecture using only 300 training samples. While the performance of both models deteriorates, the SlotFormer model predictions are now close to random predictions. On the other hand, our model is still performing better than the SlotFormer model with the full dataset available.}
    \label{tab:data_efficiency}
    \resizebox{.48\textwidth}{!}{%
    \begin{tabular}{c|c|c|c|c}
         &                         \textbf{mIoU}$\uparrow$ &                      \textbf{mIoU-FG}$\uparrow$ &                          \textbf{ARI}$\uparrow$ &                       \textbf{ARI-FG}$\uparrow$ \\
\midrule
      Ours-300 & $23.4$\textcolor{darkgray}{\scriptsize$\pm1.7$} & $19.0$\textcolor{darkgray}{\scriptsize$\pm2.0$} & $33.0$\textcolor{darkgray}{\scriptsize$\pm2.8$} & $35.5$\textcolor{darkgray}{\scriptsize$\pm5.3$} \\
Slotformer-300 & $12.1$\textcolor{darkgray}{\scriptsize$\pm0.3$} &  $5.9$\textcolor{darkgray}{\scriptsize$\pm0.3$} & $11.1$\textcolor{darkgray}{\scriptsize$\pm0.8$} &  $2.6$\textcolor{darkgray}{\scriptsize$\pm0.2$} \\\midrule
          Ours & $32.1$\textcolor{darkgray}{\scriptsize$\pm0.7$} & $29.2$\textcolor{darkgray}{\scriptsize$\pm0.8$} & $48.2$\textcolor{darkgray}{\scriptsize$\pm1.0$} & $82.9$\textcolor{darkgray}{\scriptsize$\pm2.8$} \\
    SlotFormer & $15.1$\textcolor{darkgray}{\scriptsize$\pm0.1$} &  $9.4$\textcolor{darkgray}{\scriptsize$\pm0.1$} & $16.9$\textcolor{darkgray}{\scriptsize$\pm0.3$} &  $6.1$\textcolor{darkgray}{\scriptsize$\pm0.4$} \\
    \end{tabular}}
\end{table}

\subsection{Joint Latent State}\label{sec:exp_joint}
Here we analyze whether the separation of the latent state into Gestalt and dynamics factors is necessary by working on only a single latent state without separation, without both the latent state encoder and decoder. As can be seen in Tab.~\ref{tab:original_performance}, the performance decreases significantly when not performing latent state separation. However, the performance was still above that of the SlotFormer model, indicating that even poor dynamics integration can be beneficial.

\begin{table}[tb]
    \centering
    \caption{Performance comparison of our proposed architecture (Ours) and a variant that does not separate the latent state into Gestalt and dynamics factors (Ours-Single). For reference the performance of the SlotFormer model is also shown.}
    \label{tab:original_performance}
    \resizebox{.48\textwidth}{!}{%
    \begin{tabular}{c|c|c|c|c}
         &                         \textbf{mIoU}$\uparrow$ &                      \textbf{mIoU-FG}$\uparrow$ &                          \textbf{ARI}$\uparrow$ &                       \textbf{ARI-FG}$\uparrow$ \\
\midrule
Ours-Single & $21.6$\textcolor{darkgray}{\scriptsize$\pm1.1$} & $16.7$\textcolor{darkgray}{\scriptsize$\pm1.3$} & $28.1$\textcolor{darkgray}{\scriptsize$\pm1.9$} & $26.1$\textcolor{darkgray}{\scriptsize$\pm4.6$} \\\midrule
       Ours & $32.1$\textcolor{darkgray}{\scriptsize$\pm0.7$} & $29.2$\textcolor{darkgray}{\scriptsize$\pm0.8$} & $48.2$\textcolor{darkgray}{\scriptsize$\pm1.0$} & $82.9$\textcolor{darkgray}{\scriptsize$\pm2.8$} \\
 SlotFormer & $15.1$\textcolor{darkgray}{\scriptsize$\pm0.1$} &  $9.4$\textcolor{darkgray}{\scriptsize$\pm0.1$} & $16.9$\textcolor{darkgray}{\scriptsize$\pm0.3$} &  $6.1$\textcolor{darkgray}{\scriptsize$\pm0.4$} \\
    \end{tabular}}
\end{table}

\section{Conclusion}
We have introduced a scheme to integrate procedural knowledge into deep learning models and specialized this approach for a video prediction case. We have shown that the prediction performance can be significantly improved if one uses knowledge about underlying dynamics as opposed to learning in a data-driven fashion alone. However, we also highlighted the benefit of (1) a sensible latent state separation in order to facilitate the use of the procedural knowledge, and (2) the use of a parallel prediction model that corrects the dynamics prediction and models Gestalt and dynamics interdependencies. Future work is focused on increasing the benefit further for inaccurate or incomplete knowledge integration, as this enables the use in more complex settings. Also, the current need for ground truth conditioning in the first frame limits applicability in some settings, and as such semi-supervised or even completely unsupervised state discovery increase the utility of our approach. Last, the application to video prediction downstream tasks such as MPC, VQA, or more complex system parameter estimation are all potential extensions of this work.

{\small
\bibliographystyle{ieee_fullname}
\bibliography{ControllableVideoGeneration}
}

\end{document}